\pdfoutput=1
\documentclass[11pt]{article}
\usepackage{acl}
\usepackage{times}
\usepackage{latexsym}
\usepackage[T1]{fontenc}
\usepackage[utf8]{inputenc}
\usepackage{microtype}
\usepackage{inconsolata}
\usepackage{graphicx}
\usepackage{booktabs}
\usepackage{amsmath}
\usepackage{graphicx}
\usepackage{subcaption}
\usepackage{graphicx}
\usepackage{subcaption}
\usepackage{xurl}

\title{Article and Comment Frames Shape the Quality of Online Comments}

\author{Matteo Guida \qquad Yulia Otmakhova \qquad Eduard Hovy \qquad Lea Frermann \\
  School of Computing and Information Systems,\\ The University of Melbourne \\
    \texttt{\href{mailto:guida@student.unimelb.edu.au}{guida@student.unimelb.edu.au}}, \\
    \texttt{\{y.otmakhova,eduard.hovy,lea.frermann\}@unimelb.edu.au}
}

\begin{document}
\maketitle

\begin{abstract}
Framing theory posits that how information is presented shapes audience responses, but computational work has largely 
ignored audience reactions. While recent work showed that article framing systematically shapes the {\it content} of reader responses, this paper asks: 
does framing also affect response \textit{quality}?
Analyzing 1M comments across 2.7K news articles, we operationalize quality as comment health. We find that article frames significantly predict comment health while controlling for topic, and that comments that adopt the article frame are healthier than those that depart from it. Further, unhealthy top-level comments tend to generate more unhealthy responses, independent of the frame being used in the comment. Our results establish a link between framing theory and discourse quality, laying the groundwork for downstream applications. We illustrate this potential with a pro-active frame-aware LLM-based system to mitigate unhealthy discourse.\footnote{Code and data are released in this GitHub Repo: \href{https://github.com/mattguida/healthiness-framing}{healthiness-framing}.}
\end{abstract}

\section{Introduction}

Online news platforms have associated comment threads in which readers not only can directly engage with the article, but also with each other. Maintaining constructive dialogue in these spaces, however, is challenging. While extensive computational work has focused on predicting comment toxicity or quality \citep{pavlopoulos-etal-2017-deeper, Founta_Djouvas_Chatzakou_Leontiadis_Blackburn_Stringhini_Vakali_Sirivianos_Kourtellis_2018, shankaran2024analyzingtoxicitydeepconversations}, it treats all discussions as equivalent, overlooking a fundamental insight from framing theory: the perspective (\textit{framing}) through which issues are presented shapes how audiences respond \citep{entman1993framing, scheufele1999framing, chong2007framing}.

Traditionally, computational framing research analyzed only source texts (e.g., news articles, political speeches, social media posts) while ignoring how audiences responded to framed content \citep{card_media_2015, field_framing_2018, liu2019detecting}.\footnote{See \citet{ali_survey_2022} and \citet{otmakhova_frermann_media_2024} for overviews of computational framing analysis.} Recent work has begun to address this gap by jointly analysing framing in articles and audience responses at scale \citep{guida2025retainreframecomputationalframework}. Analyzing news comment sections, they found that on average less than half of the comments retain the dominant article frame—the rest selectively adopt secondary frames or introduce entirely new ones. This proved to be true especially for value-laden frames (Morality, Fairness and Cultural Identity), suggesting that framing shapes response \textit{content}: the perspectives that people adopt when discussing issues. 
However, a critical question remains: 
if the choice of frames shape how audiences interpret and reconstruct messages, does it also influence the \textit{quality} of discussions?

We connect framing theory with discourse quality across 1M comments on 2.7K  articles from The New York Times and The Globe and Mail. We operationalize discourse quality as \textbf{comment health}: the extent to which contributions 
are made in good faith, invite engagement, and focus on substance rather than hostility \citep{price-etal-2020-six}. Healthy comments may include robust disagreement but remain constructive in tone. Health is thus different from the widely used concept {\it toxicity}, and the two have been shown to correlate poorly~\citep{price-etal-2020-six}.

\begin{table*}[t]
\centering
\small
\begin{tabular}{p{7.5cm}p{7.5cm}}
\toprule
\textbf{Healthy} & \textbf{Unhealthy} \\
\midrule
"How do you spell exploitation? This is a disgusting practice that sanctions abuses of fundamental rights and human decency. We cannot continue to condone it." & "Funny how the 'free' market is not OK in this situation. So 'not OK' that corps lobbied the 'Harper Government' to create a policy to circumvent the free market. How f@\#ked is that? The Harper govt must go" \\
\midrule
"Excellent article. For too long, the aboriginals' concerns have been treated as a problem to avoid, when it should be seen as a vital part of the whole project." & "Are you kidding? The mandatory native on all the projects I did with the Feds in northern Alberta was a uniformly drunk, indolent or absent 'partner'. What a joke" \\
\bottomrule
\end{tabular}
\caption{Examples from our corpus illustrating healthy versus unhealthy comments. Note that healthy comments may include strong language or disagreement while remaining substantive and constructive; unhealthy comments may avoid explicit profanity yet undermine discourse through dismissiveness, generalisation, or bad faith, different to the widely used concept of toxicity (See Section~\ref{sec:methods-health} and Appendix~\ref{sec:appendix-health-toxicity}).}
\label{tab:frame-health-examples}
\end{table*}

We posit two ways in which frames influence discussion quality. First, article frames may directly influence the health of comments. For instance, a news article framing immigration as a "security threat" versus an "economic opportunity" may activate different affective responses and generate more or less constructive discussions. Second, direct comments to the article serve as "secondary" framers that mediate the article's message. Healthy or unhealthy replies to such comments may unfold: a healthy top-level comment using an economic frame may elicit healthier replies than a healthy comment using a moral frame, even when both respond to the same article. We therefore ask:



\textbf{(RQ1)} Do \textbf{article frames} influence the health of \textbf{top-level comments} through (a) frame type (which frames are used) and (b) frame alignment (whether commentators match or depart from article frames)?

\textbf{(RQ2)} Does the frame of \textbf{top-level comments} influence the health of \textbf{their replies}?

We find that comment health significantly varies as a function of the article frame, when controlled for article topic.
Frame alignment also matters: comments that adopt the article frames are significantly healthier than those introducing new perspectives. Analysis of comment threads shows that healthy comments generate healthier replies through a cascade effect that operates consistently across all frame types.

We are the first to show a systematic impact of framing on discussion quality at scale on naturalistic data. Our findings can impact content moderation approaches, suggesting that 
rather than reacting to malicious content after it appears, platforms could proactively identify high-risk comments based on which frames are used, or whether commentators depart from article frames. We illustrate this through a frame, content and health aware LLM-based system that analyzes article and comments to provide real-time reformulation suggestions, helping commenters express views more constructively (see Section~\ref{ssec:system}). 
\section{Data and Methods}
\subsection{Data} Our analysis examines news articles and associated comment threads from The New York Times (NYT), a major U.S. newspaper, and The Globe and Mail (SOCC), Canada's national newspaper. We build on the dataset from \citet{guida2025retainreframecomputationalframework}, comprising news articles and comments from 2012-2018 across 11 topics (e.g., Immigration, Healthcare, Climate Change).

We extend the data set in two key ways. First, while they sampled only top-level comments, we retrieve \textit{complete comment threads} associated with these articles from the original datasets.\footnote{\citet{kolhatkar2020sfu} and \url{https://www.kaggle.com/datasets/aashita/nyt-comments}} This enables us to examine both top-level comment health (RQ1) and how health propagates through reply chains (RQ2). Second, we apply frame and health classifiers to this expanded set of replies. Table~\ref{tab:dataset-stats} summarizes the final dataset. 

\subsection{Methods}
\begin{table}[t!]
\centering
\small
\begin{tabular}{lccc}
\toprule
 & \textbf{NYT} & \textbf{SOCC} & \textbf{Total} \\
\midrule
Articles & 1,671 & 1,077 & 2,748 \\
Comments & 831.9K & 194.8K & 1.03M \\
\midrule
Depth 0 & 620.9K & 93.7K & 714.6K \\
Depth 1 & 208.5K & 69.2K & 277.7K \\
Depth 2 & 2.6K & 31.8K & 34.4K \\
Depth 3+ & 0.05K & 2.6K & 2.6K \\
\bottomrule
\end{tabular}
\caption{Dataset statistics. Depth 0: top-level comments; Depth 1: replies to top-level comments; Depth 2: nested replies. Depths 3 and beyond account for only 0.25\% of all comments.}
\label{tab:dataset-stats}
\end{table}

\paragraph{Frame Classification} We predict the primary and secondary frames of each article and comment by applying the fine-tuned RoBERTa classifier from~\citet{guida2025retainreframecomputationalframework} to assign frame labels from a taxonomy of nine generic frames, with an additional Other category, as reported in Table~\ref{tab:frame-taxonomy}. Frame predictions are obtained for all news articles and comments.

\begin{table}[h]
\centering
\small
\renewcommand{\arraystretch}{1.3}
\begin{tabular}{lp{4.5cm}}
\toprule
\textbf{Frame} & \textbf{Description} \\
\midrule
Economic & Costs, benefits, economic consequences, jobs, trade \\
Morality & Religious/ethical perspectives, moral judgments \\
Fairness \& Equality & Equal treatment, discrimination, rights, justice \\
Legality \& Crime$^*$ & Laws, constitutionality, crime, punishment \\
Political \& Policies$^+$ & Policy prescriptions, governance, partisan framing \\
Security \& Defense & National security, military, terrorism, border control \\
Health \& Safety & Public health, medical consequences, physical wellbeing \\
Cultural Identity & National identity, traditions, community belonging \\
Public Opinion & Polls, popular sentiment, public debate \\
Other & Frames not captured by the above categories \\
\bottomrule
\end{tabular}
\caption{Frame taxonomy. $^*$Merged from \textit{Legality \& jurisprudence} and \textit{Crime \& punishment}; $^+$merged from \textit{Policy prescription} and \textit{Political} \citep{card_media_2015, guida2025retainreframecomputationalframework}.}
\label{tab:frame-taxonomy}
\end{table}

\paragraph{Health Classification} 
\label{sec:methods-health}
We assign a health score to each comment following the framework of \citet{price-etal-2020-six}. We use their Unhealthy Comment Corpus (UCC) which provides binary labels indicating whether each comment \textit{has a place in a healthy conversation}, annotated by up to five crowd workers with aggregated confidence scores \citep{price-etal-2020-six}. 
Under their definition, a \textit{healthy} online conversation is one in which posts and comments exhibit observable textual characteristics conducive to constructive discourse, not overly hostile or destructive, and generally inviting engagement. Healthy conversations may include robust debate and disagreement but are typically focused on substance and ideas. 

The original data are highly imbalanced, with healthy comments comprising over 90\% of the corpus. To improve minority class representation, we resample the data by retaining only high-confidence annotations ($\geq 0.8$) and reducing majority class representation through undersampling of the healthy class. This procedure results in a more balanced dataset of approximately 10k instances (see Appendix Tables~\ref{tab:orig-splits} and~\ref{tab:balanced-splits} for further details).



\paragraph{Models} We fine-tuned two transformer-based models for  health classification: DeBERTa-v3 \citep{he2021debertav3}, ModernBERT \citep{warner2024smarterbetterfasterlonger}; and an open-source LLM; LLaMA~3.1–8B \citep{grattafiori2024llama}. DeBERTa-v3 and ModernBERT are fine-tuned using class-weighted binary cross-entropy loss to mitigate class imbalance and improve minority-class recall. For LLaMA, we apply LoRA-based instruction fine-tuning \citep{hu2022lora}.

\begin{table}[t!]
\centering
\small
\begin{tabular}{lcccc}
\toprule
\textbf{Model} & \textbf{Acc.} & \textbf{Precision} & \textbf{Recall} & \textbf{F1} \\
\midrule
LLaMA & 0.59 & 0.50 & 0.50 & 0.49 \\
ModernBERT & 0.68 & 0.69 & 0.71 & 0.67 \\
DeBERTa & 0.74 & 0.72 & 0.74 & \textbf{0.72} \\
\bottomrule
\end{tabular}
\caption{Accuracy and macro-averaged precision, recall, F1 on the re-balanced UCC test set.}
\label{tab:health-results}
\end{table}

\paragraph{Health Classification Performance}
As shown in Table~\ref{tab:health-results}, we obtained robust classifiers capable of distinguishing between healthy and unhealthy comments, with DeBERTa performing best overall. We hence use DeBERTa to predict comment health in the NYT and SOCC. 

Three authors of this paper manually annotated a subset of 100 NYT/SOCC comments with moderate inter-annotator agreement (Fleiss' $\kappa = 0.54$). The average agreement between human and model labels was 78\%, indicating reliable performance.

NYT and SOCC comment health predictions are skewed (75\% healthy). This is expected because both platforms employ manual content moderation,\footnote{Guidelines: \url{https://help.nytimes.com/hc/en-us/articles/115014792387-Comments} (NYT), \url{https://www.theglobeandmail.com/community-guidelines/} (SOCC).} which establishes a relatively high baseline for discourse quality. Despite the class imbalance, over 250k comments labelled as unhealthy are available for our main analysis



\paragraph{Health vs Toxicity} Health does not require posts and comments to be friendly, grammatically correct, well structured, or free of vulgarity, making it distinct from \textit{toxicity}. Comparing our health predictions against Perspective API toxicity scores on our corpus yields only slight agreement ($\kappa = 0.19$--$0.21$) and moderate negative correlations ($\rho \approx -0.51$, $p < 0.001$), with $20$--$24\%$ of comments we classify as unhealthy receiving low toxicity scores. Upon manual inspection, these comments are typically dismissive, use sweeping generalizations or stereotypes, or employ condescending or sarcastic tones---undermining constructive dialogue without the explicit or emotionally charged language that toxicity detectors flag. Table~\ref{tab:health-toxicity-disagreement} in the Appendix provides examples of unhealthy comments with low toxicity levels.

\section{Results}

\subsection{RQ1: Article Framing Effects on Comment Health}
We examine (1) whether article frame type predicts health, and (2) whether frame alignment (matching the article's primary frame, adopting a secondary frame, or introducing new frames) affects health, while controlling for topic. We fit mixed-effects logistic regression models to predict binary top-level comment health with random effects for article IDs and fixed effects for article topic (all models), article framing (RQ1.1) or frame alignment (RQ1.2).\footnote{Full model specifications in Appendix~\ref{sec:appendix-rq1}.} 


\paragraph{RQ1.1 Article Frame Effects}

The primary frame of the article exerts a significant influence on comment health in both outlets (NYT: $\chi^2(9) = 368.27$, $p < 0.001$; SOCC: $\chi^2(9) = 26.97$, $p = 0.001$).
Health and Economic frames consistently elicit the healthiest comments across both platforms (84--87\% healthy comments), while Political, Fairness, and Morality frames generate the least healthy discourse (72--77\% healthy), indicating that value-laden frames provoke more contentious user engagement.

\begin{figure}
\centering
\includegraphics[width=\linewidth]{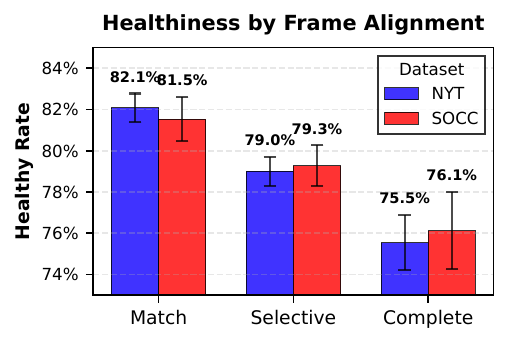}
\caption{Health by frame alignment across platforms. A clear gradient emerges where comments primary matching article frames are the healthiest, followed by selective reframing (adopting secondary frames present in the article), with complete reframing (introducing frames absent from the article) showing the lowest health. Error bars represent 95\% confidence intervals.}
\label{fig:frame-alignment-health}
\end{figure}

\paragraph{RQ1.2 Frame Alignment Effects}
The degree of alignment between article and comment frames significantly predicts comment health on both platforms (NYT: $\chi^2(2) = 340.61$, $p < .001$; SOCC: $\chi^2(2) = 48.16$, $p < .001$), after controlling for topic. A clear gradient emerges (Figure~\ref{fig:frame-alignment-health}): comments that adopt the primary article frame are the healthiest, followed by selective reframing (adopting secondary frames present in the article), with complete reframing (introducing frames absent from the article) exhibiting the lowest health. All three pairwise comparisons between frame alignment conditions are significant at $p \ll .001$ (See Table~\ref{tab:rq1-alignment} in the Appendix).
This trend holds across topics (see Figure~\ref{fig:frame-alignment-by-topic} in Appendix~\ref{sec:appendix-frame-alignment-topics}).

The selective reframing category is particularly informative.
Readers who remain within the article's \textit{frame repertoire}, even when shifting away from its primary emphasis, tend to engage more constructively than those who introduce completely novel perspectives.

\paragraph{Topic Effects}

Unsurprisingly, article topic alone also strongly predicts comment health when controlling for frame type (NYT: $\chi^2(10) = 371.85$, $p < .001$; SOCC: $\chi^2(10) = 435.52$, $p < .001$). Some topics (e.g., Health, Education) consistently foster more constructive discourse, while others are associated with substantially lower health (e.g., Trump). Full results by topic are in Appendix Figure~\ref{fig:frame-alignment-by-topic}.



\subsection{RQ2: Comment Framing Effects on Reply Health}
We model mean reply health by fitting a linear regression with top-level comment health, top-level comment frame and topic as well as  health$\times$frame interactions as predictors. We then test (1) how top-level comment health impacts subsequent discussion and (2) whether this trend differs across frames; while again controlling for topic.\footnote{Full modeling details in Appendix~\ref{sec:appendix-rq2}.} 

Top-level comment health strongly predicts subsequent reply health in both outlets (NYT: $\beta = 0.095$, $SE = 0.018$, $t = 5.38$, $p < 0.001$; SOCC: $\beta = 0.126$, $SE = 0.023$, $t = 5.40$, $p < 0.001$),  in line with similar studies on toxicity in Reddit threads~\citep{shankaran2024analyzingtoxicitydeepconversations}.
Topic effects persist and are significant in both datasets ($p < 0.001$), with Education and Healthcare topics associated with healthier reply chains across platforms. By contrast, top-level comment frame has only modest and inconsistent main effects on reply health: in NYT, Political-frame top comments show slightly lower reply health than the Cultural baseline ($\beta = -0.034$, $p = 0.041$), while in SOCC, Economic-frame comments show slightly higher reply health ($\beta = 0.070$, $p = 0.013$) and Fairness-frame comments slightly lower ($\beta = -0.102$, $p = 0.012$); no other frame contrasts reach significance in either platform. Crucially, none of the Health~$\times$~Frame interaction terms are significant in either platform (all $p > 0.10$). This trend does not significantly vary by top-level comment frame. In other words, healthy top comments initiate healthy discussions and unhealthy top comments initiate unhealthy discussions -- largely independent of their topic or frame. 
Full results are in Appendix Table~\ref{tab:rq2-regression-combined}.\\
\\
Combined with RQ1, where frame type and alignment significantly predicted baseline comment health, our findings suggest a \textbf{two-stage model} of how framing shapes online discourse. In the first stage, article framing sets the \textit{discourse environment}: which frames an article uses, and whether commenters align with or depart from those frames, significantly predicts the health of top-level comments (RQ1). On average, Health and Economic frames foster healthier discussions; Political and Fairness frames generate less healthy ones; and complete reframing is consistently associated with the lowest health. In the second stage, once top-level comments are posted, their \textit{health} is the primary driver of reply quality (RQ2). A healthy Economic-frame comment and a healthy Morality-frame comment generate healthy replies at comparable rates; similarly for unhealthy comments. In other words: \textit{framing determines the health of the first response; health then determines what that response generates.} This two-stage structure suggests that proactive moderation interventions are best applied early, at the article-framing or top-level-comment stage, before health cascades through reply chains.


\subsection{Frame-Aware Content Moderation}
\label{ssec:system}
Current content moderation systems operate reactively: they detect and remove toxic content after posting. This approach addresses symptoms rather than causes, it penalizes users without helping them improve, and cannot prevent unhealthy discussions before they escalate.

Our findings suggest an alternative: {proactive, frame-aware moderation}.
Two levers identified in our results make this actionable. First, article frame type is a \textit{leading indicator} of discourse risk: articles framed around Morality, Fairness, or Politics reliably attract less healthy top-level comments, while Health and Economic frames attract healthier ones (RQ1.1). Platforms and editors could use this signal \textit{before} comments arrive to calibrate moderation intensity or prime readers with context-setting prompts. Second, frame alignment at the comment level provides an early, post-submission signal: comments that depart entirely from the article's frame are significantly more likely to be unhealthy (OR $\approx 1.33$--$1.35$ compared to frame-matching comments; RQ1.2). Since RQ2 shows that top-level comment health cascades through reply threads, catching and reformulating unhealthy or misaligned top-level comments early can interrupt downstream deterioration before it begins.

Together, these two signals---article frame risk and comment frame alignment---make it possible to stratify comments by likely discourse impact rather than simply flagging surface-level toxicity after the fact.
We illustrate this through a prototype system that analyzes article framing, comment framing, their frame alignment 
and comment health. Based on these inputs, the system stratifies comments into three risk levels (high, medium and low), and suggests LLM-based reformulations informed by full context. The system can be accessed
\href{https://mpprng--comment-moderation-agent-commentmoderationservice-serve.modal.run/}{online at this \textbf{link}} (see Appendix~\ref{sec:system-details} for complete technical specifications).



\section{Conclusion}

Our findings establish that framing and discourse health are intertwined: article frames significantly predict comment health (value-laden frames like Morality and 
Fairness generating the least healthy discourse), frame alignment matters (complete reframing showing the strongest negative effects), and comment health cascades uniformly through reply chains regardless of frame type (RQ2).

We provide empirical evidence for framing theory \citep{neuman1992common, scheufele1999framing}: audiences do not passively absorb frames but actively 
reconstruct them, and these reconstruction patterns predict not just \textit{what} they say but \textit{how} they engage. Our findings carry practical value, 
illustrated through a pro-active, frame-aware moderation prototype using frame classification and health detection to suggest constructive reformulations.




\section*{Limitations}
Our analysis examines correlational patterns rather than causal mechanisms. Our binary health operationalization enables large-scale analysis while simplifying the multidimensional nature of discourse quality. We analyze two major English-language news outlets (NYT, Globe and Mail) from 2012-2018, which may reflect outlet-specific moderation practices and temporal context. The frame and health classifiers achieve robust performance on our datasets, though generalization to other platforms and languages remains to be validated.

\section*{Ethical Considerations}
This study was approved by the Human Ethics Committee (Reference No.\ 2025-32561-72301-5) and has been conducted according to the corresponding ethical standards.

Our moderation prototype is intended to support constructive discourse, not suppress dissent. We caution against using frame alignment as a sole criterion for moderation, as novel perspectives that depart from article framing may represent valuable contributions. The operationalization of "health" reflects norms from specific cultural contexts and should be adapted for other communities.

\section*{Acknowledgments}
This paper was written with the support of the Melbourne Research Scholarship provided by the University of Melbourne to MG. This work was also supported by the Australian Research Council Discovery Early Career Research Award (Grant No.\ DE230100761).

\label{sec:bibtex}
\bibliography{custom}

@inproceedings{ali_survey_2022,
    title = "{A Survey of Computational Framing Analysis Approaches}",
    author = "Ali, Mohammad  and
      Hassan, Naeemul",
    editor = "Goldberg, Yoav  and
      Kozareva, Zornitsa  and
      Zhang, Yue",
    booktitle = "Proceedings of the 2022 Conference on Empirical Methods in Natural Language Processing",
    month = dec,
    year = "2022",
    address = "Abu Dhabi, United Arab Emirates",
    publisher = "Association for Computational Linguistics",
    url = "https://aclanthology.org/2022.emnlp-main.633/",
    doi = "10.18653/v1/2022.emnlp-main.633",
    pages = "9335--9348"
}

@inproceedings{card_media_2015,
	address = {Beijing, China},
	title = "{The Media Frames Corpus: Annotations of Frames across Issues}",
	shorttitle = {The media frames corpus},
	url = {https://aclanthology.org/P15-2072},
	doi = {10.3115/v1/P15-2072},
	urldate = {2024-07-04},
	booktitle = {Proceedings of the 53rd Annual Meeting of the Association for Computational Linguistics and the 7th International Joint Conference on Natural Language Processing (Volume 2: Short Papers)},
	publisher = {Association for Computational Linguistics},
	author = {Card, Dallas and Boydstun, Amber E. and Gross, Justin H. and Resnik, Philip and Smith, Noah A.},
	editor = {Zong, Chengqing and Strube, Michael},
	month = jul,
	year = {2015},
	pages = {438--444},
}

@article{chong2007framing,
author = {Chong, Dennis and Druckman, James},
year = {2007},
month = {12},
pages = {},
title = "{Framing Theory}",
volume = {10},
journal = {Annual Review of Political Science},
doi = {10.1146/annurev.polisci.10.072805.103054}
}

@article{entman1993framing,
author = {Entman, Robert},
year = {1993},
month = {12},
pages = {51-58},
title = {Framing: Toward clarification of a fractured paradigm},
volume = {43},
journal = {The Journal of Communication},
doi = {10.1111/j.1460-2466.1993.tb01304.x}
}

@inproceedings{field_framing_2018,
	address = {Brussels, Belgium},
	title = "{Framing and Agenda-setting in {Russian} News: A Computational Analysis of Intricate Political Strategies}",
	shorttitle = {Framing and agenda-setting in {Russian} news},
	url = {https://aclanthology.org/D18-1393},
	doi = {10.18653/v1/D18-1393},
	urldate = {2024-04-12},
	booktitle = {Proceedings of the 2018 Conference on Empirical Methods in Natural Language Processing},
	publisher = {Association for Computational Linguistics},
	author = {Field, Anjalie and Kliger, Doron and Wintner, Shuly and Pan, Jennifer and Jurafsky, Dan and Tsvetkov, Yulia},
	editor = {Riloff, Ellen and Chiang, David and Hockenmaier, Julia and Tsujii, Jun'ichi},
	month = oct,
	year = {2018},
	pages = {3570--3580},
}

@article{grattafiori2024llama,
  title="{The {LLaMA} 3 Herd of Models}",
  author={Grattafiori, Aaron and Dubey, Abhimanyu and Jauhri, Abhinav and Pandey, Abhinav and Kadian, Abhishek and Al-Dahle, Ahmad and Letman, Aiesha and Mathur, Akhil and Schelten, Alan and Vaughan, Alex and others},
  journal={arXiv preprint arXiv:2407.21783},
  year={2024}
}

@article{hu2022lora,
  title="{LoRA: Low-rank Adaptation of Large Language Models}",
  author={Hu, Edward J and Shen, Yelong and Wallis, Phillip and Allen-Zhu, Zeyuan and Li, Yuanzhi and Wang, Shean and Wang, Lu and Chen, Weizhu and others},
  journal={ICLR},
  volume={1},
  number={2},
  pages={3},
  year={2022}
}

@article{kolhatkar2020sfu,
  author       = {Veselin Kolhatkar and Haofen Wu and Liane Cavasso and Maite Taboada},
  title        = "{The SFU Opinion and Comments Corpus: A Corpus for the Analysis of Online News Comments}",
  journal      = {Corpus Pragmatics},
  volume       = {4},
  number       = {2},
  pages        = {155--190},
  year         = {2020},
  doi          = {10.1007/s41701-019-00065-w},
  url          = {https://doi.org/10.1007/s41701-019-00065-w},
  note         = {Published: 2 November 2019; Issue Date: June 2020}
}

@inproceedings{liu2019detecting,
    title = "{Detecting Frames in News Headlines and its Application to Analyzing News Framing Trends Surrounding {U.S.} Gun Violence}",
    author = "Liu, Siyi  and
      Guo, Lei  and
      Mays, Kate  and
      Betke, Margrit  and
      Wijaya, Derry Tanti",
    editor = "Bansal, Mohit  and
      Villavicencio, Aline",
    booktitle = "Proceedings of the 23rd Conference on Computational Natural Language Learning (CoNLL)",
    month = nov,
    year = "2019",
    address = "Hong Kong, China",
    publisher = "Association for Computational Linguistics",
    url = "https://aclanthology.org/K19-1047/",
    doi = "10.18653/v1/K19-1047",
    pages = "504--514",
}

@inproceedings{otmakhova_frermann_media_2024,
    title = "{Media Framing: A Typology and Survey of Computational Approaches across Disciplines}",
    author = "Otmakhova, Yulia  and
      Khanehzar, Shima  and
      Frermann, Lea",
    editor = "Ku, Lun-Wei  and
      Martins, Andre  and
      Srikumar, Vivek",
    booktitle = "Proceedings of the 62nd Annual Meeting of the Association for Computational Linguistics (Volume 1: Long Papers)",
    month = aug,
    year = "2024",
    address = "Bangkok, Thailand",
    publisher = "Association for Computational Linguistics",
    url = "https://aclanthology.org/2024.acl-long.822/",
    doi = "10.18653/v1/2024.acl-long.822",
    pages = "15407--15428"
}

@book{neuman1992common,
  author    = {W. Russell Neuman and Marion R. Just and Ann N. Crigler},
  title     = {Common Knowledge: News and the Construction of Political Meaning},
  publisher = {University of Chicago Press},
  year      = {1992},
  address   = {Chicago}
}

@article{scheufele1999framing,
  author       = {Dietram A. Scheufele},
  title        = "{Framing as a Theory of Media Effects}",
  journal      = {Journal of Communication},
  volume       = {49},
  number       = {1},
  pages        = {103--122},
  year         = {1999},
  doi          = {10.1111/j.1460-2466.1999.tb02784.x},
  url          = {https://doi.org/10.1111/j.1460-2466.1999.tb02784.x}
}

@inproceedings{price-etal-2020-six,
    title = "{Six Attributes of Unhealthy Conversations}",
    author = "Price, Ilan  and
      Gifford-Moore, Jordan  and
      Flemming, Jory  and
      Musker, Saul  and
      Roichman, Maayan  and
      Sylvain, Guillaume  and
      Thain, Nithum  and
      Dixon, Lucas  and
      Sorensen, Jeffrey",
    editor = "Akiwowo, Seyi  and
      Vidgen, Bertie  and
      Prabhakaran, Vinodkumar  and
      Waseem, Zeerak",
    booktitle = "Proceedings of the Fourth Workshop on Online Abuse and Harms",
    month = nov,
    year = "2020",
    address = "Online",
    publisher = "Association for Computational Linguistics",
    url = "https://aclanthology.org/2020.alw-1.15/",
    doi = "10.18653/v1/2020.alw-1.15",
    pages = "114--124"
}

@inproceedings{pavlopoulos-etal-2017-deeper,
    title = "{Deeper Attention to Abusive User Content Moderation}",
    author = "Pavlopoulos, John  and
      Malakasiotis, Prodromos  and
      Androutsopoulos, Ion",
    editor = "Palmer, Martha  and
      Hwa, Rebecca  and
      Riedel, Sebastian",
    booktitle = "Proceedings of the 2017 Conference on Empirical Methods in Natural Language Processing",
    month = sep,
    year = "2017",
    address = "Copenhagen, Denmark",
    publisher = "Association for Computational Linguistics",
    url = "https://aclanthology.org/D17-1117/",
    doi = "10.18653/v1/D17-1117",
    pages = "1125--1135"
}

@article{Founta_Djouvas_Chatzakou_Leontiadis_Blackburn_Stringhini_Vakali_Sirivianos_Kourtellis_2018, title="{Large Scale Crowdsourcing and Characterization of Twitter Abusive Behavior}", volume={12}, url={https://ojs.aaai.org/index.php/ICWSM/article/view/14991}, DOI={10.1609/icwsm.v12i1.14991}, number={1}, journal={Proceedings of the International AAAI Conference on Web and Social Media}, author={Founta, Antigoni and Djouvas, Constantinos and Chatzakou, Despoina and Leontiadis, Ilias and Blackburn, Jeremy and Stringhini, Gianluca and Vakali, Athena and Sirivianos, Michael and Kourtellis, Nicolas}, year={2018}, month={Jun.} }

@misc{he2021debertav3,
      title="{DeBERTaV3: Improving DeBERTa using ELECTRA-Style Pre-Training with Gradient-Disentangled Embedding Sharing}", 
      author={Pengcheng He and Jianfeng Gao and Weizhu Chen},
      year={2021},
      eprint={2111.09543},
      archivePrefix={arXiv},
      primaryClass={cs.CL}
}

@inproceedings{warner2024smarterbetterfasterlonger,
    title = "Smarter, Better, Faster, Longer: A Modern Bidirectional Encoder for Fast, Memory Efficient, and Long Context Finetuning and Inference",
    author = {Warner, Benjamin  and
      Chaffin, Antoine  and
      Clavi{\'e}, Benjamin  and
      Weller, Orion  and
      Hallstr{\"o}m, Oskar  and
      Taghadouini, Said  and
      Gallagher, Alexis  and
      Biswas, Raja  and
      Ladhak, Faisal  and
      Aarsen, Tom  and
      Adams, Griffin Thomas  and
      Howard, Jeremy  and
      Poli, Iacopo},
    editor = "Che, Wanxiang  and
      Nabende, Joyce  and
      Shutova, Ekaterina  and
      Pilehvar, Mohammad Taher",
    booktitle = "Proceedings of the 63rd Annual Meeting of the Association for Computational Linguistics (Volume 1: Long Papers)",
    month = jul,
    year = "2025",
    address = "Vienna, Austria",
    publisher = "Association for Computational Linguistics",
    url = "https://aclanthology.org/2025.acl-long.127/",
    doi = "10.18653/v1/2025.acl-long.127",
    pages = "2526--2547",
    ISBN = "979-8-89176-251-0"
}

@misc{guida2025retainreframecomputationalframework,
      title="{Retain or Reframe? A Computational Framework for the Analysis of Framing in News Articles and Reader Comments}", 
      author={Matteo Guida and Yulia Otmakhova and Eduard Hovy and Lea Frermann},
      year={2025},
      eprint={2507.04612},
      archivePrefix={arXiv},
      primaryClass={cs.CL},
      url={https://arxiv.org/abs/2507.04612}, 
}

@misc{shankaran2024analyzingtoxicitydeepconversations,
      title="{Analyzing Toxicity in Deep Conversations: A Reddit Case Study}", 
      author={Vigneshwaran Shankaran and Rajesh Sharma},
      year={2024},
      eprint={2404.07879},
      archivePrefix={arXiv},
      primaryClass={cs.CL},
      url={https://arxiv.org/abs/2404.07879}, 
}

\appendix

\section{UCC Data Splits}
The original train / val / test UCC data splits and our label-balanced data splits are shown in Table~\ref{tab:orig-splits} and \ref{tab:balanced-splits}, respectively.

\begin{table}[h]
\centering
\small
\begin{tabular}{lccc}
\toprule
& Healthy & Unhealthy & Total \\
\midrule
Train & 32{,}848 & 2{,}655 & 35{,}503 \\
Val & 4{,}091 & 336 & 4{,}427 \\
Test & 4{,}105 & 320 & 4{,}425 \\
\midrule
Total & 41{,}044 & 3{,}311 & 44{,}355 \\
\bottomrule
\end{tabular}
\caption{Original class distribution in UCC splits.}
\label{tab:orig-splits}
\end{table}

\begin{table}[h]
\centering
\small
\begin{tabular}{lccc}
\toprule
& Healthy & Unhealthy & Total \\
\midrule
Train & 5{,}298 & 2{,}649 & 7{,}947 \\
Val & 662 & 331 & 993 \\
Test & 662 & 331 & 993 \\
\midrule
Total & 6{,}622 & 3{,}311 & 9{,}933 \\
\bottomrule
\end{tabular}
\caption{Balanced high-confidence split used for fine-tuning.}
\label{tab:balanced-splits}
\end{table}

\section{Health and Toxicity}
\label{sec:appendix-health-toxicity}

\citet{price-etal-2020-six} showed that `health' and `toxicity' only have a weak correspondence in UCC. We verify this trend on our NYT and SOCC corpus, adding support for our decision to focus on comment health, specifically.

To assess how our healthy discourse classifier relates to toxicity detection, we compared our health predictions with Perspective API toxicity scores on the SOCC and NYT datasets. We binarized toxicity scores at a threshold of $0.5$ and calculated Cohen's $\kappa$ and correlations. 

Results show slight agreement ($\kappa = 0.19$--$0.21$) with moderate negative correlations ($\rho \approx -0.51$, $p < 0.001$). $20$--$24\%$ of comments classified as unhealthy have low toxicity scores, accounting for $96$--$97\%$ of instances classified as unhealthy, but with low toxicity scores. Upon manual investigation of a subset of disagreements, we found that these comments are dismissive, unproductive, use sweeping generalizations or stereotypes, and employ condescending or sarcastic tones. 
As such, these comments undermine constructive and healthy dialogue, however, they do not necessarily use explicit or overly  emotional language which would give rise to a toxic label. Examples of such cases are provided in Table~\ref{tab:health-toxicity-disagreement}.

\begin{table*}[h]
\centering
\small
\begin{tabular}{p{9cm}ccp{3cm}}
\toprule
\textbf{Comment Text} & \textbf{Healthy} & \textbf{Toxicity} & \textbf{Characteristics} \\
\midrule
"A few general comments: did not read the column. Stopped reading Lady Astor long ago; interesting she did not editorialize about Chris Spence, admitted serial plagiarizer, head of TDSB; Really, who still reads her columns? A columnist who's plagiarizes?" & 0 & 0.127 & Dismissive, unconstructive \\
\addlinespace
"The number of lies, distortions and exaggerations in this article are many. Wonder where she got them?" & 0 & 0.243 & Accusatory \\
\addlinespace
"America: it's an amazing country full of a lot of fine people. But like wayward teenagers you can't tell them anything. They know what's best, even when it's so obvious that their behavior is harming themselves and others. There will be no significant change in America's gun-cult mentality. Not now, nor even if there's 20 more shootings like Newtown. Guns are fetish objects to Americans, and they're worshiped accordingly. Either accept that, or live and visit elsewhere." & 0 & 0.188 & Stereotypes, generalization \\
\addlinespace
"Get with the program, Margaret, or you'll be sent to a re-education camp. It's 'climate change.'" & 0 & 0.049 & Sarcasm \\
\bottomrule
\end{tabular}
\caption{Examples of comments classified as unhealthy (Healthy=0) but with low Perspective API toxicity scores (<0.5).}
\label{tab:health-toxicity-disagreement}
\end{table*}

\section{Illustrative Examples of Reframing}
\label{sec:appendix-reframing-examples}

To illustrate the patterns observed in our analysis, we present representative examples of selective and complete reframing from our corpus in Table~\ref{tab:selective-reframing-examples} and Table~\ref{tab:complete-reframing-examples}.

\begin{table*}[h]
\centering
\footnotesize
\begin{tabular}{p{2.5cm}p{1.2cm}p{1.5cm}p{1.8cm}p{5.0cm}p{1.3cm}}
\toprule
\textbf{Headline} & \textbf{Topic} & \textbf{Article Frame} & \textbf{Secondary Frames} & \textbf{Comment} & \textbf{Comment Frame} \\
\midrule
Without Obamacare, I Will Get Sicker, Faster & Healthcare & Health & Economic, Political, Morality & My granddaughter once asked me the difference between Democrats and Republicans... 'Democrats,' I said, 'want to help people. Republicans don't want to help people.' & Political \\
\midrule
The B.C. teachers' case gets a failing grade & Education & Political and Policies & Economic, Legality and Crime & Gotta love how the second paragraph of this piece is worded in such a way that it suggests that the BC government's attempts to illegally strip teachers of their constitutional right to bargain class size and composition actually puts it in a better light than the BCTF. & Legality and Crime \\
\bottomrule
\end{tabular}
\caption{Examples of selective reframing where commenters adopt secondary frames present in the article.}
\label{tab:selective-reframing-examples}
\end{table*}

\begin{table*}[h]
\centering
\footnotesize
\begin{tabular}{p{2.5cm}p{1.2cm}p{1.5cm}p{1.8cm}p{5.0cm}p{1.3cm}}
\toprule
\textbf{Headline} & \textbf{Topic} & \textbf{Article Frame} & \textbf{Secondary Frames} & \textbf{Comment} & \textbf{Comment Frame} \\
\midrule
Trump Is Insulting Our Intelligence & Trump & Political and Policies & Legality and Crime, Morality, Cultural Identity & We keep discussing \#45 as though he were sane. He is not... He has no true friends and his family don't care... It would be tragic if he wasn't President, but it's simply horrifying now. & Health and Safety \\
\midrule
Mr. Trump, Meet My Family & Immigration & Cultural Identity & Legality and Crime, Morality, Political and Policies, Security and Defense, Health and Safety, Fairness and Equality & I wonder why the folks who voted for Trump are not encouraged to move where there are jobs and retrain... Immigrants cost money from taxpayers, so why not use that money to train current Americans... & Economic \\
\bottomrule
\end{tabular}
\caption{Examples of complete reframing where commenters introduce frames entirely absent from the article.}
\label{tab:complete-reframing-examples}
\end{table*}

\section{Overall Topic Health}
\label{sec:appendix-topic-health}

Table~\ref{tab:topic-health-overall} details the rate of healthy comments by topic. The variation in health by topic is independent of framing and highly significant in both datasets (NYT: $\chi^2(10) = 371.9$, $p < .001$; SOCC: $\chi^2(10) = 435.5$, $p < .001$). Healthcare and Education consistently foster the most constructive discourse, while Trump coverage generates the lowest health rates.

\begin{table}[ht]
\centering
\small
\begin{tabular}{@{}lcc@{}}
\toprule
\textbf{Topic} & \textbf{NYT Health} & \textbf{SOCC Health} \\
\midrule
Healthcare & 88\% & 95\% \\
Education & 88\% & 87\% \\
Climate Change & 85\% & 86\% \\
Abortion & 83\% & 86\% \\
Syria & 83\% & 77\% \\
Gun Control & 79\% & 78\% \\
Israel & 79\% & 70\% \\
Russia & 78\% & 72\% \\
Gay Rights & 78\% & 76\% \\
Immigration & 78\% & 80\% \\
Trump & 68\% & 67\% \\
\bottomrule
\end{tabular}
\caption{Mean comment health by topic, sorted by NYT rate. Differences between topics are statistically significant ($p < 0.001$) for both platforms.}
\label{tab:topic-health-overall}
\end{table}

\subsection{Frame Alignment Effects by Topic}
\label{sec:appendix-frame-alignment-topics}

\begin{figure*}[h]
\centering
\includegraphics[width=\linewidth]{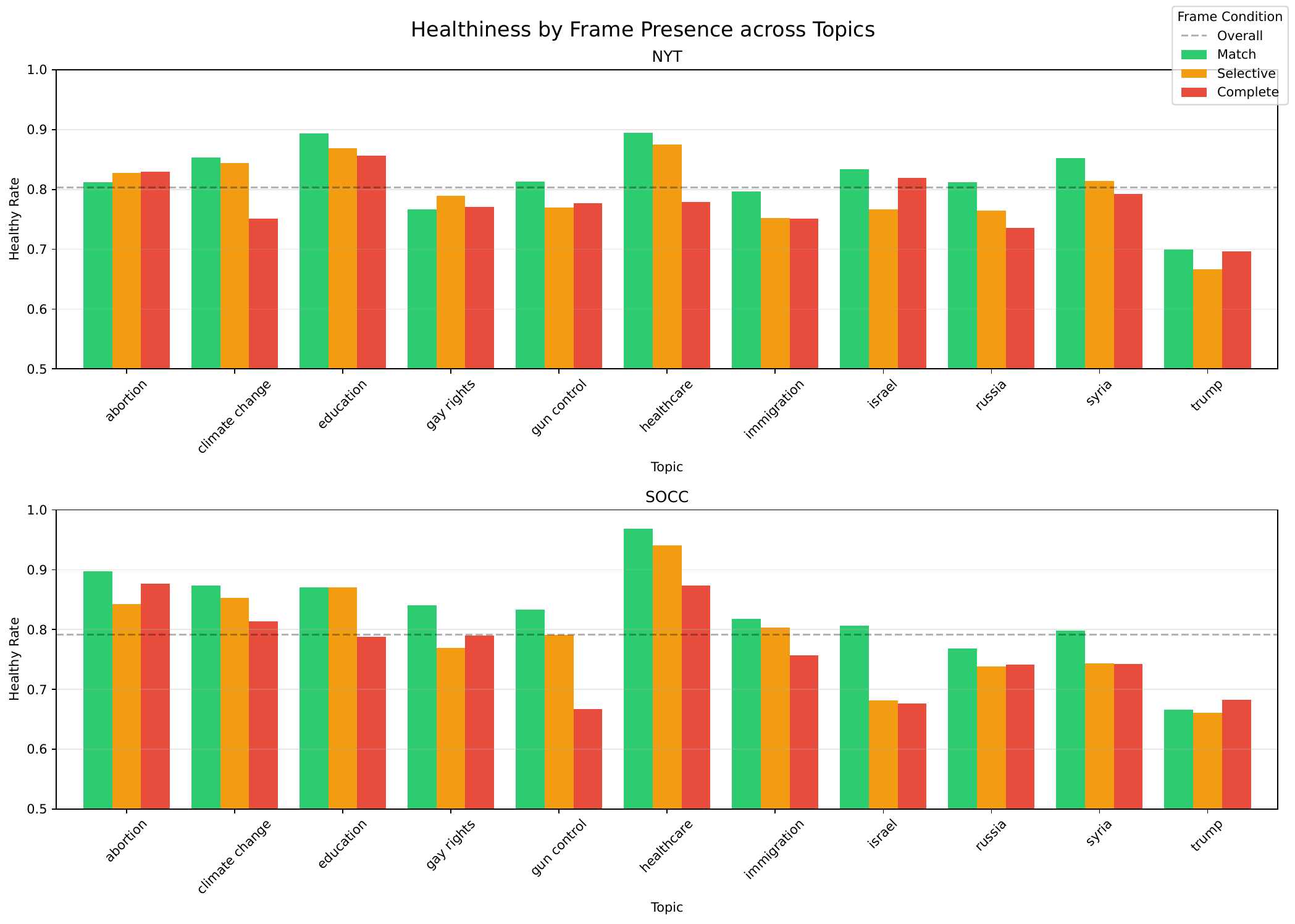}
\caption{Comment health by frame alignment (Match, Diff in Article, Never in Article) across topics for NYT (top) and SOCC (bottom). The dashed line represents overall health.}
\label{fig:frame-alignment-by-topic}
\end{figure*}

Figure~\ref{fig:frame-alignment-by-topic} shows how frame alignment effects vary across topics. The consistent pattern---where matching frames yield healthier comments than selective or complete reframing---holds across most topics.

\section{Regression Analyses and Full Results}
\subsection{RQ1: Article effects on top-level comments}
\label{sec:appendix-rq1}

For RQ1, we ask how article framing influences comment health. First, we examine whether article frame type predicts health. We fit a mixed-effects logistic regression model that predicts (binary) health of a top-level comment based on article frame, topic as fixed effects and article ID as random effect due to multiple measurements (top comments) for the same article. Our variable of interest is article\_frame, and we include topic as a control variable:
\begin{align}
    \text{health} \sim \text{article\_frame} + \text{topic} + (1|\text{id}).
\end{align}

Second, we test the effect of frame alignment between articles and comments predicts health (frame\_condition). Similarly, we fit a logistic regression mixed-effects model replacing article\_frame with frame\_condition,
\begin{align}
    \text{health} \sim \text{frame\_condition} + \text{topic} + (1|\text{id}),
\end{align}
where frame\_condition is a three-level factor indicating whether comments match the article's primary frame, selectively reframe by adopting a secondary frame present in the article, or completely reframe by introducing frames not in the article.

\paragraph{Full Results} Tables~\ref{tab:rq1-frame-alignment-regression} and \ref{tab:rq1-article-frame-regression} present the full regression models for frame alignment and article frame effects, respectively. Tables~\ref{tab:rq1-alignment} and Table~\ref{tab:rq1-article-frames} summarize the estimated marginal means.

\begin{table*}[h]
\centering
\small
\begin{tabular}{lrrrr|rrrr}
\toprule
& \multicolumn{4}{c}{\textbf{NYT}} & \multicolumn{4}{c}{\textbf{SOCC}} \\
\cmidrule(lr){2-5} \cmidrule(lr){6-9}
\textbf{Predictor} & \textbf{Est.} & \textbf{SE} & \textbf{$z$} & \textbf{$p$} & \textbf{Est.} & \textbf{SE} & \textbf{$z$} & \textbf{$p$} \\
\midrule
(Intercept) & 1.727 & 0.099 & 17.37 & $<0.001$ & 1.862 & 0.173 & 10.77 & $<0.001$ \\
\midrule
\multicolumn{9}{l}{\textit{Frame Alignment (vs Match)}} \\
Selective & $-0.108$ & 0.009 & $-11.94$ & $<0.001$*** & $-0.132$ & 0.029 & $-4.51$ & $<0.001$*** \\
Complete & $-0.298$ & 0.018 & $-16.31$ & $<0.001$*** & $-0.289$ & 0.044 & $-6.62$ & $<0.001$*** \\
\midrule
\multicolumn{9}{l}{\textit{Article Topic (vs Abortion)}} \\
Climate Change & $-0.022$ & 0.113 & $-0.20$ & 0.843 & 0.047 & 0.176 & 0.27 & 0.788 \\
Education & 0.542 & 0.131 & 4.14 & $<0.001$*** & 0.131 & 0.185 & 0.71 & 0.480 \\
Gay Rights & $-0.328$ & 0.179 & $-1.83$ & 0.067 & $-0.268$ & 0.219 & $-1.22$ & 0.221 \\
Gun Control & $-0.288$ & 0.117 & $-2.45$ & 0.014* & $-0.500$ & 0.214 & $-2.34$ & 0.020* \\
Healthcare & 0.410 & 0.107 & 3.84 & $<0.001$*** & 0.976 & 0.209 & 4.67 & $<0.001$*** \\
Immigration & $-0.380$ & 0.111 & $-3.42$ & $<0.001$*** & $-0.316$ & 0.181 & $-1.74$ & 0.081 \\
Israel & $-0.265$ & 0.131 & $-2.02$ & 0.043* & $-0.792$ & 0.191 & $-4.15$ & $<0.001$*** \\
Russia & $-0.374$ & 0.105 & $-3.55$ & $<0.001$*** & $-0.744$ & 0.186 & $-3.99$ & $<0.001$*** \\
Syria & $-0.095$ & 0.137 & $-0.69$ & 0.491 & $-0.428$ & 0.181 & $-2.37$ & 0.018* \\
Trump & $-0.862$ & 0.107 & $-8.09$ & $<0.001$*** & $-1.103$ & 0.177 & $-6.24$ & $<0.001$*** \\
\midrule
\multicolumn{9}{l}{\textbf{Random Effects}} \\
\multicolumn{9}{l}{Article ID (Intercept) Variance: 0.369 (NYT), 0.201 (SOCC)} \\
\multicolumn{9}{l}{Article ID (Intercept) SD: 0.608 (NYT), 0.448 (SOCC)} \\
\midrule
\multicolumn{9}{l}{\textbf{Model Fit}} \\
\multicolumn{5}{l}{NYT: AIC = 363749.3, BIC = 363899.9} & \multicolumn{4}{l}{SOCC: AIC = 39157.4, BIC = 39277.5} \\
\midrule
\multicolumn{9}{l}{\textbf{Overall Effects (Type II Wald $\chi^2$ tests)}} \\
\multicolumn{5}{l}{Frame Alignment: $\chi^2(2) = 340.61$, $p < 0.001$***} & \multicolumn{4}{l}{Frame Alignment: $\chi^2(2) = 48.16$, $p < 0.001$***} \\
\multicolumn{5}{l}{Topic: $\chi^2(10) = 627.65$, $p < 0.001$***} & \multicolumn{4}{l}{Topic: $\chi^2(10) = 662.46$, $p < 0.001$***} \\
\bottomrule
\end{tabular}
\caption{Mixed-effects logistic regression predicting comment health from frame alignment, NYT and SOCC.}
\label{tab:rq1-frame-alignment-regression}
\end{table*}

\begin{table*}[h]
\centering
\small
\begin{tabular}{lrrrr|rrrr}
\toprule
& \multicolumn{4}{c}{\textbf{NYT}} & \multicolumn{4}{c}{\textbf{SOCC}} \\
\cmidrule(lr){2-5} \cmidrule(lr){6-9}
\textbf{Predictor} & \textbf{Est.} & \textbf{SE} & \textbf{$z$} & \textbf{$p$} & \textbf{Est.} & \textbf{SE} & \textbf{$z$} & \textbf{$p$} \\
\midrule
(Intercept) & 1.608 & 0.108 & 14.90 & $<0.001$ & 1.734 & 0.196 & 8.85 & $<0.001$ \\
\midrule
\multicolumn{9}{l}{\textit{Article Frame (vs Cultural)}} \\
Economic & $-0.104$ & 0.098 & $-1.06$ & 0.288 & 0.185 & 0.097 & 1.92 & 0.055 \\
Fairness & $-0.560$ & 0.275 & $-2.04$ & 0.042* & $-0.544$ & 0.195 & $-2.79$ & 0.005** \\
Health & 0.390 & 0.085 & 4.61 & $<0.001$*** & 0.051 & 0.122 & 0.42 & 0.673 \\
Legality & $-0.262$ & 0.085 & $-3.06$ & 0.002** & $-0.144$ & 0.133 & $-1.08$ & 0.279 \\
Morality & 0.016 & 0.147 & 0.11 & 0.914 & $-0.125$ & 0.216 & $-0.58$ & 0.563 \\
Opinion & $-0.213$ & 0.157 & $-1.36$ & 0.174 & 0.017 & 0.206 & 0.08 & 0.936 \\
Other & 0.096 & 0.083 & 1.16 & 0.245 & $-0.032$ & 0.097 & $-0.33$ & 0.743 \\
Political & $-0.553$ & 0.069 & $-7.96$ & $<0.001$*** & $-0.064$ & 0.087 & $-0.73$ & 0.466 \\
Security & $-0.219$ & 0.125 & $-1.75$ & 0.080 & $-0.015$ & 0.113 & $-0.14$ & 0.891 \\
\midrule
\multicolumn{9}{l}{\textit{Article Topic (vs Abortion)}} \\
Climate Change & 0.167 & 0.103 & 1.62 & 0.104 & 0.024 & 0.184 & 0.13 & 0.898 \\
Education & 0.801 & 0.121 & 6.59 & $<0.001$*** & 0.177 & 0.193 & 0.92 & 0.360 \\
Gay Rights & $-0.051$ & 0.165 & $-0.31$ & 0.758 & $-0.155$ & 0.225 & $-0.69$ & 0.491 \\
Gun Control & $-0.033$ & 0.108 & $-0.30$ & 0.762 & $-0.433$ & 0.219 & $-1.98$ & 0.048* \\
Healthcare & 0.547 & 0.096 & 5.70 & $<0.001$*** & 0.906 & 0.211 & 4.30 & $<0.001$*** \\
Immigration & 0.041 & 0.103 & 0.40 & 0.687 & $-0.314$ & 0.185 & $-1.70$ & 0.089 \\
Israel & 0.023 & 0.122 & 0.19 & 0.847 & $-0.736$ & 0.198 & $-3.71$ & $<0.001$*** \\
Russia & 0.104 & 0.098 & 1.06 & 0.291 & $-0.715$ & 0.194 & $-3.68$ & $<0.001$*** \\
Syria & 0.075 & 0.138 & 0.55 & 0.584 & $-0.384$ & 0.190 & $-2.03$ & 0.043* \\
Trump & $-0.385$ & 0.099 & $-3.90$ & $<0.001$*** & $-0.992$ & 0.185 & $-5.35$ & $<0.001$*** \\
\midrule
\multicolumn{9}{l}{\textbf{Random Effects}} \\
\multicolumn{9}{l}{Article ID (Intercept) Variance: 0.292 (NYT), 0.195 (SOCC)} \\
\multicolumn{9}{l}{Article ID (Intercept) SD: 0.540 (NYT), 0.442 (SOCC)} \\
\midrule
\multicolumn{9}{l}{\textbf{Model Fit}} \\
\multicolumn{5}{l}{NYT: AIC = 363757.2, BIC = 363983.1} & \multicolumn{4}{l}{SOCC: AIC = 39192.5, BIC = 39372.7} \\
\midrule
\multicolumn{9}{l}{\textbf{Overall Effects (Type II Wald $\chi^2$ tests)}} \\
\multicolumn{5}{l}{Article Frame: $\chi^2(9) = 368.27$, $p < 0.001$***} & \multicolumn{4}{l}{Article Frame: $\chi^2(9) = 26.97$, $p = 0.001$**} \\
\multicolumn{5}{l}{Topic: $\chi^2(10) = 371.85$, $p < 0.001$***} & \multicolumn{4}{l}{Topic: $\chi^2(10) = 435.52$, $p < 0.001$***} \\
\bottomrule
\end{tabular}
\caption{Mixed-effects logistic regression predicting comment health from article frames, NYT and SOCC.}
\label{tab:rq1-article-frame-regression}
\end{table*}

\begin{table*}[h]
\centering
\small
\begin{minipage}{0.48\linewidth}
    \centering
    \begin{tabular}{lcc}
    \toprule
    \textbf{Alignment} & \textbf{NYT} & \textbf{SOCC} \\
    \midrule
    Match & 82.9\% & 83.1\% \\
    Selective & 81.3\% & 81.1\% \\
    Complete & 78.2\% & 78.6\% \\
    \midrule
    \multicolumn{3}{l}{\textit{Pairwise Comparisons (Tukey-adjusted):}} \\
    Match vs Selective & OR = 1.11*** & OR = 1.14*** \\
    Match vs Complete & OR = 1.35*** & OR = 1.33*** \\
    Selective vs Complete & OR = 1.21*** & OR = 1.17*** \\
    \bottomrule
    \end{tabular}
    \caption{Estimated marginal means: Frame alignment effects on health. Matching is the reference level.}
    \label{tab:rq1-alignment}
    \vspace{0.5em}
    
    \footnotesize{Note: Results averaged over topic levels. All pairwise comparisons significant at $p < 0.001$.}
\end{minipage}
\hfill
\begin{minipage}{0.48\linewidth}
    \centering
    \begin{tabular}{lcc}
    \toprule
    \textbf{Article Frame} & \textbf{NYT} & \textbf{SOCC} \\
    \midrule
    Health & 89.2\% & 82.4\% \\
    Other & 86.1\% & 81.2\% \\
    Cultural & 84.9\% & 81.7\% \\
    Economic & 83.5\% & 84.3\% \\
    Opinion & 81.9\% & 81.9\% \\
    Security & 81.8\% & 81.5\% \\
    Legality & 81.2\% & 79.4\% \\
    Morality & 85.1\% & 79.7\% \\
    Political & 76.3\% & 80.7\% \\
    Fairness & 76.2\% & 72.1\% \\
    \bottomrule
    \end{tabular}
    \caption{Estimated marginal means: Comment health by article frame.}
    \label{tab:rq1-article-frames}
    \vspace{0.5em}
    
    \footnotesize{Note: Results averaged over topic levels. Frames sorted by NYT health rate.}
\end{minipage}
\end{table*}

\subsection{RQ2: Top-level comment effects on reply thread health}
\label{sec:appendix-rq2}

For RQ2, we model mean reply health (MRH) using ordinary least squares (OLS) linear regression, specifying the model as follows:
\begin{equation}
\begin{aligned}
\text{MRH} \sim\;&
\text{top\_c\_health} + \text{top\_c\_frame} \\
 &+ \text{article\_topic}\\
 &+ \text{top\_c\_health} \times \text{top\_c\_frame},
\end{aligned}
\end{equation}
where top\_c is short for top\_comment. We include article\_frame again as a control factor and also capture the interaction of health and frame in the interaction term.

\paragraph{Full Results} Table~\ref{tab:rq2-regression-combined} present full coefficient estimates, standard errors, $t$-statistics, and $p$-values for both platforms.

\begin{table*}[h]
\centering
\footnotesize
\begin{tabular}{lrrrr|rrrr}
\toprule
& \multicolumn{4}{c}{\textbf{NYT}} & \multicolumn{4}{c}{\textbf{SOCC}} \\
\cmidrule(lr){2-5} \cmidrule(lr){6-9}
\textbf{Predictor} & \textbf{$\beta$} & \textbf{SE} & \textbf{$t$} & \textbf{$p$} & \textbf{$\beta$} & \textbf{SE} & \textbf{$t$} & \textbf{$p$} \\
\midrule
(Intercept) & 0.706 & 0.019 & 37.39 & $<0.001$ & 0.633 & 0.029 & 21.49 & $<0.001$ \\
\midrule
\multicolumn{9}{l}{\textit{Top Comment Health}} \\
Healthy (vs Unhealthy) & 0.095 & 0.018 & 5.38 & $<0.001$*** & 0.126 & 0.023 & 5.40 & $<0.001$*** \\
\midrule
\multicolumn{9}{l}{\textit{Top Comment Frame (vs Cultural)}} \\
Economic & 0.026 & 0.023 & 1.14 & 0.256 & 0.070 & 0.028 & 2.48 & 0.013* \\
Fairness & $-0.003$ & 0.031 & $-0.09$ & 0.932 & $-0.102$ & 0.041 & $-2.51$ & 0.012* \\
Health & 0.006 & 0.023 & 0.27 & 0.787 & 0.047 & 0.044 & 1.07 & 0.286 \\
Legality & 0.020 & 0.021 & 0.99 & 0.323 & $-0.022$ & 0.048 & $-0.47$ & 0.642 \\
Morality & $-0.034$ & 0.020 & $-1.75$ & 0.079 & $-0.046$ & 0.031 & $-1.48$ & 0.140 \\
Opinion & $-0.032$ & 0.044 & $-0.74$ & 0.458 & $-0.023$ & 0.048 & $-0.49$ & 0.626 \\
Other & $-0.020$ & 0.017 & $-1.15$ & 0.251 & $-0.019$ & 0.022 & $-0.83$ & 0.406 \\
Political & $-0.034$ & 0.016 & $-2.04$ & 0.041* & $-0.023$ & 0.023 & $-1.03$ & 0.303 \\
Security & 0.057 & 0.045 & 1.26 & 0.207 & $-0.053$ & 0.047 & $-1.13$ & 0.258 \\
\midrule
\multicolumn{9}{l}{\textit{Article Topic (vs Abortion)}} \\
Climate Change & 0.030 & 0.011 & 2.81 & 0.005** & 0.053 & 0.021 & 2.48 & 0.013* \\
Education & 0.058 & 0.012 & 4.75 & $<0.001$*** & 0.087 & 0.022 & 3.91 & $<0.001$*** \\
Gay Rights & 0.026 & 0.016 & 1.61 & 0.108 & 0.094 & 0.027 & 3.51 & $<0.001$*** \\
Gun Control & 0.028 & 0.012 & 2.46 & 0.014* & 0.060 & 0.027 & 2.24 & 0.025* \\
Healthcare & 0.034 & 0.010 & 3.30 & 0.001** & 0.122 & 0.024 & 5.12 & $<0.001$*** \\
Immigration & 0.017 & 0.011 & 1.62 & 0.105 & 0.063 & 0.022 & 2.87 & 0.004** \\
Israel & 0.042 & 0.013 & 3.21 & 0.001** & $-0.025$ & 0.024 & $-1.06$ & 0.289 \\
Russia & 0.010 & 0.010 & 0.99 & 0.320 & 0.006 & 0.023 & 0.24 & 0.807 \\
Syria & 0.030 & 0.014 & 2.17 & 0.030* & 0.038 & 0.022 & 1.72 & 0.085 \\
Trump & $-0.024$ & 0.010 & $-2.29$ & 0.022* & $-0.004$ & 0.022 & $-0.18$ & 0.855 \\
\midrule
\multicolumn{9}{l}{\textit{Interactions: Health $\times$ Frame}} \\
Healthy $\times$ Economic & 0.003 & 0.024 & 0.13 & 0.894 & $-0.044$ & 0.030 & $-1.44$ & 0.151 \\
Healthy $\times$ Fairness & $-0.024$ & 0.035 & $-0.70$ & 0.487 & 0.024 & 0.047 & 0.50 & 0.614 \\
Healthy $\times$ Health & 0.039 & 0.025 & 1.56 & 0.118 & $-0.013$ & 0.047 & $-0.29$ & 0.776 \\
Healthy $\times$ Legality & $-0.006$ & 0.022 & $-0.28$ & 0.782 & 0.003 & 0.052 & 0.07 & 0.948 \\
Healthy $\times$ Morality & $-0.012$ & 0.022 & $-0.54$ & 0.589 & $-0.006$ & 0.037 & $-0.15$ & 0.879 \\
Healthy $\times$ Opinion & 0.039 & 0.046 & 0.83 & 0.406 & 0.001 & 0.053 & 0.02 & 0.988 \\
Healthy $\times$ Other & 0.014 & 0.019 & 0.74 & 0.457 & $-0.004$ & 0.025 & $-0.15$ & 0.878 \\
Healthy $\times$ Political & $-0.010$ & 0.018 & $-0.58$ & 0.563 & $-0.009$ & 0.025 & $-0.35$ & 0.727 \\
Healthy $\times$ Security & $-0.054$ & 0.047 & $-1.14$ & 0.254 & 0.048 & 0.050 & 0.95 & 0.340 \\
\midrule
\multicolumn{9}{l}{\textbf{Model Fit}} \\
\multicolumn{9}{l}{NYT: $R^2 = 0.029$; Adj. $R^2 = 0.028$; $F(29, 72{,}800) = 73.82$, $p < 0.001$; Residual SE: 0.342} \\
\multicolumn{9}{l}{SOCC: $R^2 = 0.050$; Adj. $R^2 = 0.049$; $F(29, 20{,}013) = 36.36$, $p < 0.001$; Residual SE: 0.334} \\
\bottomrule
\end{tabular}
\caption{Linear regression predicting mean reply health from top comment health, frame, and topic, NYT and SOCC.}
\label{tab:rq2-regression-combined}
\end{table*}

\section{Frame-Aware Moderation System: Technical Specification}
\label{sec:system-details}

This section provides complete technical details of the frame-aware moderation prototype described in Section~\ref{ssec:system}, accessible at the following link: \url{https://mpprng--comment-moderation-agent-commentmoderationservice-serve.modal.run/}.

The prototype system is deployed on Modal\footnote{\url{https://modal.com}}, a serverless GPU infrastructure platform. This deployment approach offers cost efficiency for research prototypes but introduces cold-start latency when the system has been idle.

\paragraph{Cold Start Behavior}
When no requests have been made for approximately 5--10 minutes, the system enters an idle state. The first subsequent request triggers initialization of:
\begin{itemize}
    \item Ollama server and Gemma 3:1b model loading
    \item DeBERTa-v3-base healthiness classifier
    \item RoBERTa-based frame classifier
\end{itemize}
Our deployment makes use of NVIDIA A10G GPU with 24GB VRAM.

\subsection{System Architecture}

Given a user-based keyword input (e.g., ``Climate change''), our system scrapes the three most recent articles from the The Conversation website\footnote{\url{https://theconversation.com/us/}, CC BY-ND 4.0} and classifies the article frames, the comment frames, the type of reframing (or match) and the comment health. These components, along with the article and comment text, are parsed by an LLM, which is then prompted to make suggestions based on decision heuristics. Table~\ref{tab:system-architecture} provides a complete overview of the system pipeline.

\begin{table*}[htb]
\centering
\small
\setlength{\tabcolsep}{6pt}

\begin{tabular}{p{3cm}p{5cm}p{6cm}}
\toprule
\textbf{Component} & \textbf{Input / Function} & \textbf{Implementation Details} \\
\midrule
\multicolumn{3}{l}{\textbf{Article Analysis}} \\
\midrule
Text Processing & Full article text & Sentence tokenization (NLTK \texttt{punkt}) \\
Frame Classification & Sentence-level frame detection & Fine-tuned RoBERTa model \citep{guida2025retainreframecomputationalframework} (10-frame taxonomy) \\
Output & Sentence-level frame labels & Each sentence annotated as \texttt{[\{frame, confidence\}]} \\
\midrule
\multicolumn{3}{l}{\textbf{Comment Analysis}} \\
\midrule
Text Processing & Comment text & Sentence tokenization (NLTK \texttt{punkt}) \\
Frame Classification & Sentence-level frame detection & Same RoBERTa model as article analysis \\
Health Prediction & Comment-level quality assessment & Fine-tuned DeBERTa-v3-base (Binary: healthy / unhealthy) \\
Frame Alignment & Article--comment frame comparison & Primary-frame overlap and divergence measures \\
\midrule
\multicolumn{3}{l}{\textbf{Risk Assessment}} \\
\midrule
Risk Stratification & Health + alignment signals & Rule-based aggregation (Table~\ref{tab:risk-rules}) \\
Moderation Decision & Risk level & Boolean allow / block decision \\
\midrule
\multicolumn{3}{l}{\textbf{Intervention Generation}} \\
\midrule
Context Construction & Inputs to LLM & Article text, top-5 article frames, comment text, comment frames, alignment status, health score \\
LLM Inference & Reformulation generation & Gemma 3:1b via Ollama \\
Output Format & Moderation guidance & Structured JSON: \texttt{\{risk\_level, suggestions[], allow\_post\}} \\
\bottomrule
\end{tabular}
\caption{System architecture for frame-aware comment moderation. Articles and comments are analyzed separately for framing and health, combined into a risk assessment, and used to generate context-aware interventions via an LLM.}
\label{tab:system-architecture}

\vspace{1em}

\begin{tabular}{p{2cm}p{2.5cm}p{2.5cm}p{1.5cm}}
\toprule
\textbf{Risk Level} & \textbf{Health} & \textbf{Alignment} & \textbf{Action} \\
\midrule
High & $< 0.3$ & Any & Suggest + Flag \\
High & $< 0.5$ & Complete & Suggest + Flag \\
\midrule
Medium & $[0.3, 0.6)$ & Any & Suggest \\
Medium & $\geq 0.6$ & Selective/Complete & Suggest \\
\midrule
Low & $\geq 0.6$ & Match & Allow \\
\bottomrule
\end{tabular}
\caption{Risk stratification decision rules.}
\label{tab:risk-rules}

\end{table*}

\subsection{LLM Prompt Structure}

For reformulation suggestions, we use Gemma 3:1b deployed via Ollama for local inference. The prompt includes full context (article text, frames, comment, alignment) and requests structured JSON output with risk level confirmation, 2--3 specific suggestions, and an allow/block recommendation. Example prompt structure is shown in Figure~\ref{fig:llm-prompt}.

\begin{figure}[h]
\centering
\fbox{
    \begin{minipage}{0.95\columnwidth}
        \small\ttfamily
        \textbf{System Instruction:} \\
        You are an AI comment moderator. Analyze this comment for health and frame transfer (reframing). Provide constructive suggestions only when the comment is unhealthy or uses a completely different perspective from the article.
        
        \vspace{0.5em}
        \textbf{CONTEXT:} \\
        \{context\}
        
        \vspace{0.5em}
        \textbf{Article Text:} \{article\}...
        
        \vspace{0.5em}
        \textbf{Comment to Analyze:} \{comment\}
        
        \vspace{0.5em}
        \textbf{Trigger:} This comment requires intervention due to: \{intervention\_trigger\}
        
        \vspace{0.5em}
        \textbf{Task:} \\
        Based on health and frame transfer analysis: \\
        1. Confirm the risk level (low, medium, high). \\
        2. Provide 2--3 specific, constructive reformulations that:
        \begin{itemize}
            \setlength\itemsep{0em}
            \item Improve health if unhealthy
            \item Help align comment with article frames if reframing is detected
            \item Maintain the core message
        \end{itemize}
        3. Determine if the original comment should be allowed.
        
        \vspace{0.5em}
        Provide a JSON response.
    \end{minipage}
}
\caption{LLM prompt used for comment reformulation.}
\label{fig:llm-prompt}
\end{figure}

\subsection{Interface Example}

\begin{figure*}[t]
    \centering
    \includegraphics[width=0.85\textwidth]{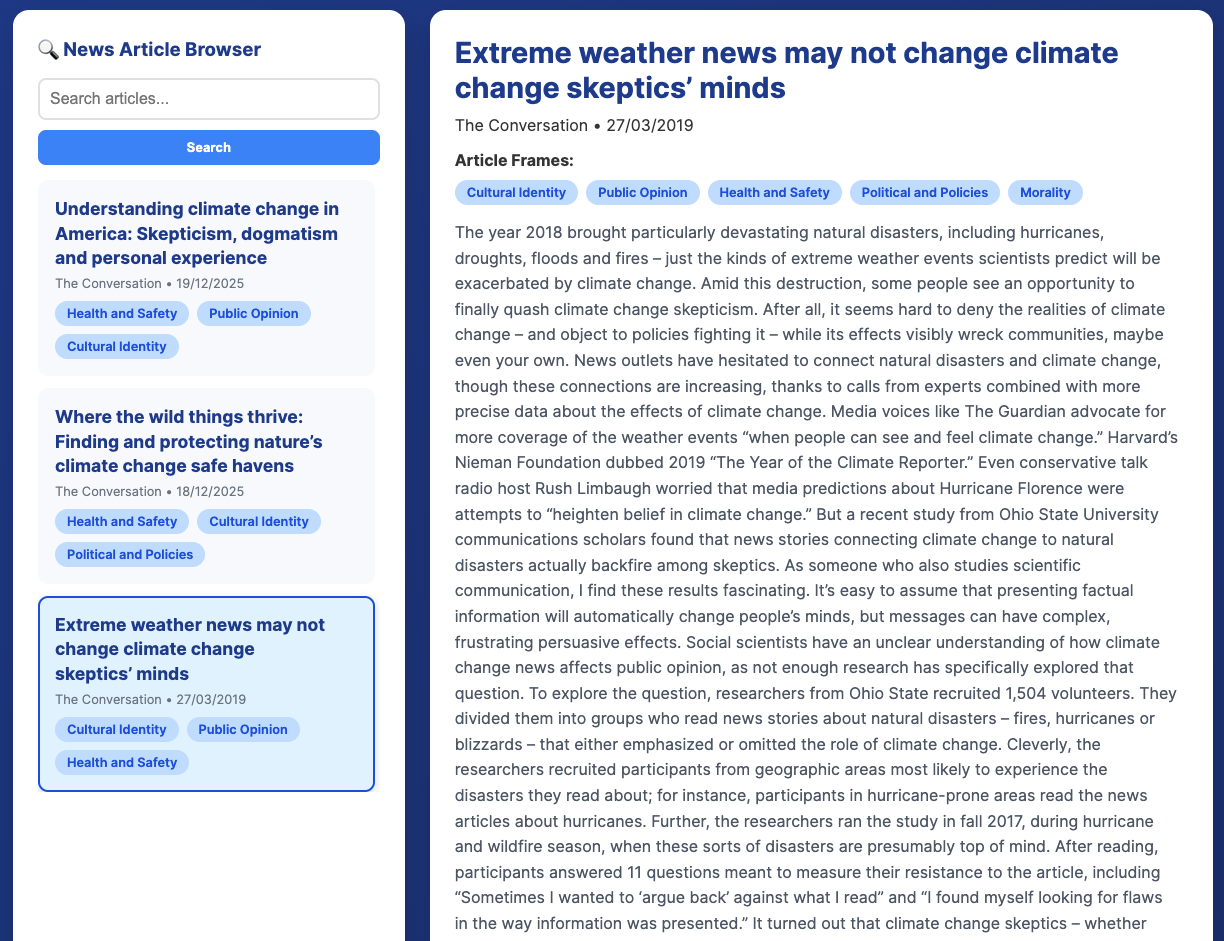}
    \caption{Landing page with topic search functionality. Users can enter keywords or select predefined topics to retrieve 3 relevant articles from The Conversation. Article view with sentence-level frame analysis. Detected frames are displayed as tags; hovering over a frame highlights corresponding sentences in the article text.}
    \label{fig:interface-search}
\end{figure*}

\begin{figure*}[t]
    \centering
    \includegraphics[width=0.85\textwidth]{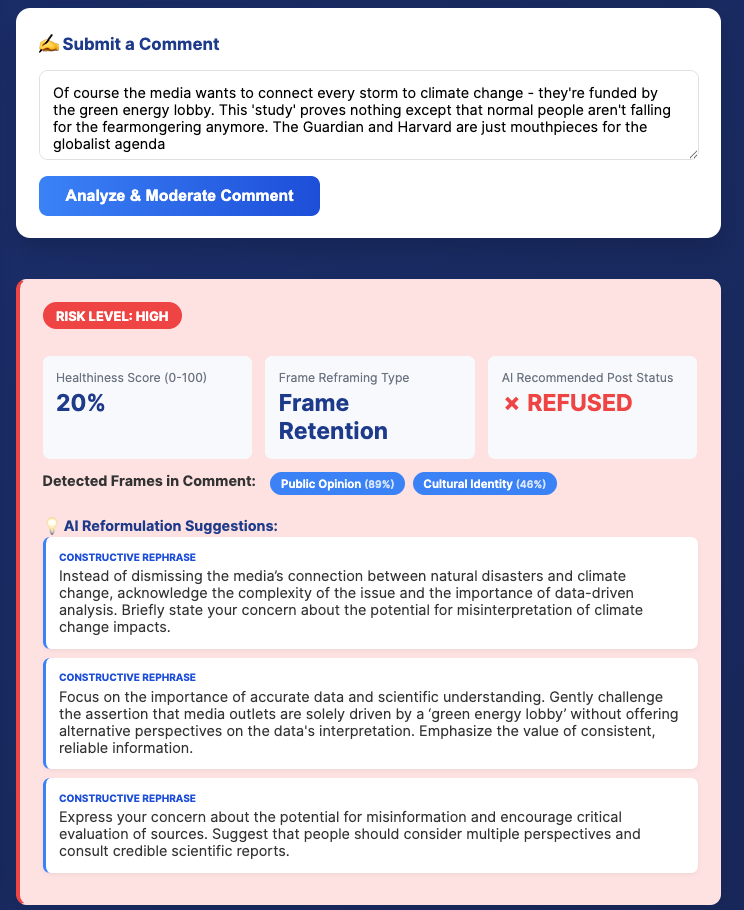}
    \caption{Moderation output for an unhealthy comment expressing climate skepticism. The system assigns a high risk level based on healthiness score, detects frames and retention, and recommends refusing the comment. The LLM generates three ways to reformulate more constructively while preserving the user's core concern and idea.}
    \label{fig:interface-unhealthy}
    \end{figure*}

Figures~\ref{fig:interface-search} and~\ref{fig:interface-unhealthy} illustrate the system interface and moderation workflow, with an example of comment analysis on a news article.

\clearpage

\end{document}